\documentclass{ecai}
\usepackage{times}
\usepackage{graphicx}
\usepackage{latexsym}
\usepackage{amsmath}

\begin{document}

\title{Deep Density-aware Count Regressor}

\author{Zhuojun Chen\institute{Baidu Inc, China, email: georgechenzj@outlook.com} \and Junhao Cheng\institute{Baidu Inc, China, email: ponmma@whu.edu.cn}  
	\and Yuchen Yuan\institute{Baidu Inc, China, email: yuanyuchen02@baidu.com} 
	\\
	\and Dongping Liao\institute{Beihang University, China, email: ldpbuaa@gmail.com}
	\and Yizhou Li\institute{Tokyo Institute of Technology, Japan, email: yli@ok.sc.e.titech.ac.jp} \and Jiancheng Lv\textsuperscript{*}\institute{Sichuan University, China, email: 
	lvjiancheng@scu.edu.cn, \textsuperscript{*} corresponding author}}

\maketitle
\bibliographystyle{ecai}

\begin{abstract}
We seek to improve crowd counting as we perceive limits of currently prevalent density map estimation approach on both prediction accuracy and time efficiency. 
We leverage multilevel pixelation of density map as it helps improve SNR of training data and therefore, reduce prediction error.
To achieve a better model, we introduce multilayer gradient fusion for 
training a density-aware global count regressor. More specifically, on training stage, a backbone network receives gradients from multiple branches to learn the density 
information, whereas those branches are to be detached to accelerate inference. By taking advantages of such method, our model improves benchmark results on 
public datasets and exhibits itself to be a new solution to crowd counting problems in practice. Our code is publicly available at:

https://github.com/GeorgeChenZJ/deepcount
\end{abstract}

\section{Introduction}
%

Crowd counting is a task to count people in image. It is mainly used in real-life for automated public monitoring such as surveillance and traffic control. Different from 
object detection, crowd counting aims at recognizing arbitrarily sized targets in various situations including sparse and cluttering scenes at the same time. Figure 1 illustrates 
some of those challenging scenarios. In recent years, crowd counting has drawn more attention from computer vision researchers and has been in significant progress.

Early methods \cite{ge2009marked,idrees2015detecting,viola2005detecting,gall2011hough,li2008estimating,zhao2008segmentation} attempt to solve the problem by detecting every individual pedestrian in the crowd. These methods often perform poorly in the face of complex conditions such as those illustrated. The recent development of crowd counting comes from DNN-based methods which have achieved commendable performance. These 
methods \cite{zhang2016single,babu2017switching,sindagi2017generating,zhang2015cross,cao2018scale,ranjan2018iterative,liu2018crowd,sam2018top,shi2018crowd} concentrate on generating the demanding density maps before integrating them to the count. They are therefore categorised into density map-based methods. However, density maps have yet in effect to show too much importance in practice except for opportunely providing for demonstration, but are expensive 
to compute, and their quality is difficult to guarantee. Meanwhile, methods that regress the global count directly have remained untouched for a while in research frontiers.

Raising state-of-the-art performances in many works, density maps have shown undeniable contribution to the improvement of prediction precision. One advantage of being 
density map-based may be that information with respect to location, scale alike is fed to the network through density supervision. Consequently, multi-scale or multi-column 
architectures \cite{zhang2016single,babu2017switching,sindagi2017generating,cao2018scale,ranjan2018iterative} are usually adopted to fuse features from different scales to capture these kinds of information. Still, there exists two main drawbacks: first, computational cost drastically increases along with the growth of number of columns; second, useful information learned by low-level detectors might be lost through forward propagation. Likewise, 
supervision information contained in gradients would be attenuated through backward propagation, making low-level detectors difficult to learn.

\begin{figure}[t]
\centerline{\includegraphics[scale=0.5]{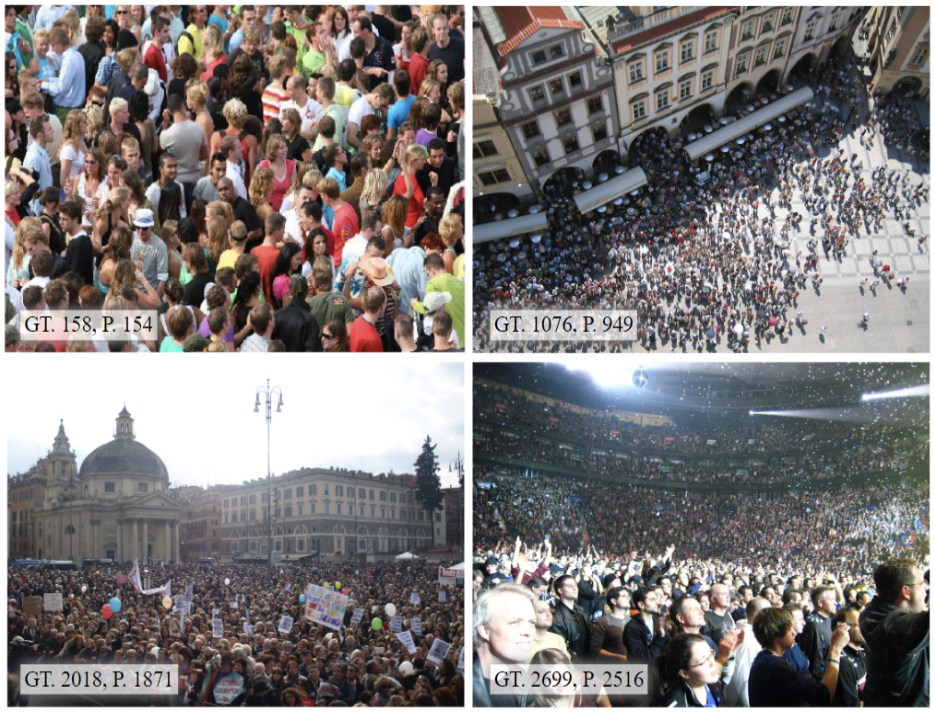}}
\caption{Representative images for challenges of non-uniform density, intra- and inter-scene variations in scale and perspective, and cluttering. In the figure, GT. means ground truth 
and P. means prediction made by our model. } \label{figure1}
\end{figure}

Besides, noise locally occurs due to uncertainty of annotation position on large-scale targets while Gaussian dispersion can only alleviate the issue to short extend. On the contrary, 
a global count is immune to local noise, but does not carry location information which can be crucial for the network to develop attention on elusive targets.

To address these problems, we propose a novel Gradient Fusion based model called DeepCount for crowd counting (network architecture shown in Figure 2), making efforts to both 
avert expenditure of multi-column architecture and enhance resistance to local annotation noise. As in Figure 2, our proposed model contains a backbone network with convolution 
layers deeply regressing a global count. Some auxiliary modules branch out to produce density maps with corresponding spatial dimensions and to feed gradients back to the backbone. 
There are five branches having different depths and independent parameters so as to learn features in different aspects. Each branch will directly access different levels of the backbone to 
inculcate knowledge to it deeply and make it more perceptive on the density distribution of the image, namely to be density-aware.

In inference phase, the backbone network can be used unaccompanied by any branches as a regressor to solely predict the global count at fastest speed, or, if needed, with 
an auxiliary branch to also visualise a density map. Expensive computation is taken out, but with functionality promised.

Compared to other multi-column methods, our model fuses gradients other than features and avoids relying on the noisily supervised and computationally expensive density maps to 
make prediction. By so doing, our model incorporates advantages of accuracy, flexibility, and efficiency.

Extensive experimental results on four benchmark datasets demonstrate significant improvements of our method against the state-of-the-art methods on Shanghai Tech Part A, Part B and 
UCF-QNRF datasets and excellent performance on Mall dataset.

The rest of the paper is structured as follow: we review literatures for crowd counting in section 2; section 3 provides the detailed interpretation of our method; section 4 reports experiment results; 
in section 5, we further discuss our findings and insights; the paper is to be concluded in section 6.

\section{Related works}
\subsection{Detection-based methods}
Early crowd counting methods tend to rely on detection approach. Low-level hand-crafted features such as Histograms of Oriented Gradients, silhouette-oriented features are exploited for traditional classifiers 
such as Support Vector Machine and Random Forest \cite{ge2009marked,viola2005detecting,gall2011hough,li2008estimating,zhao2008segmentation}. Following are CNN-based methods (e.g. Faster R-CNN \cite{ren2015faster}) which have shown credible precision \cite{liu2018decidenet}. Nonetheless, in such times when the subject of crowd counting was more on the stage of pedestrian detection, performances of these methods on highly dense crowd scenes were similarly limited. 

\subsection{Count regression-based methods}
Count regression-based methods are proposed to overcome limits encountered by detection-based methods. The idea of these methods is to regress a global count from the input image. There are methods using 
ridge regression \cite{chen2012feature,chan2008privacy}, log-linear regression \cite{ma2013crossing} or MLP \cite{kong2006viewpoint} on low-level hand-crafted features to estimate the count. While they work satisfactorily on invariant scenes of sparse density, hand-crafted features can hardly represent enough variance and intricacy in complex counting scenarios. Alternatively, with the development of deep learning, features can be black-boxed and deeply learned to target the goal. Early success of 
applying deep learning methods on crowd counting would be the end-to-end deep CNN regression model by Wang et al. \cite{wang2015deep}. Though, deep learning methods quickly narrowed onto density map-based methods which 
have prevailed over the years since, and it was not until recently in \cite{idrees2018composition} that Idrees et al. reported excellent experiment results on global count regression by advanced CNNs: Resnet101 \cite{he2016deep} and Densenet201 \cite{huang2017densely}, notwithstanding their focus on density map estimation.

\subsection{Density map-based methods}
Rodriguez et al. \cite{rodriguez2011density} first suggest the use of density map can improve crowd counting results significantly. It is supported by Zhang et al. \cite{zhang2015cross} whose model produces small density map patches as well as the patch count 
at its last layer. Following this density map approach, Zhang et al. \cite{zhang2016single} propose a multi-column architecture (MCNN) to also address scale variance of the targets. Inspired by such, Cao et al. \cite{cao2018scale} introduce Scale Aggregation 
Network (SANet) which aggregates multi-scale features and fuses them in every layer. Likewise, Switching-CNN \cite{babu2017switching} has independent columns of CNN similar to multi-column network with different receptive fields, and ic-CNN \cite{ranjan2018iterative} aims at predicting high-resolution density maps with two branches. Another set of methods devote themselves to trace context information as well as other abstractions all in a bit to improve the predicted density maps \cite{sindagi2017generating,liu2018crowd,sam2018top,shi2018crowd}. On the other hand, CSRNet \cite{li2018csrnet} builds dilated convolution layers upon a VGG-16 \cite{simonyan2014very} backbone straightforward without too many manoeuvres, yet it reports excellent results and therefore becomes more practiced at present.

Differently, our method embodies heterogeneity of multi-column methods and straightforwardness of CSRNet whilst appearing as an existence that is both regression-based and density map-based.

\begin{figure}[t]
\centerline{\includegraphics[scale=0.4]{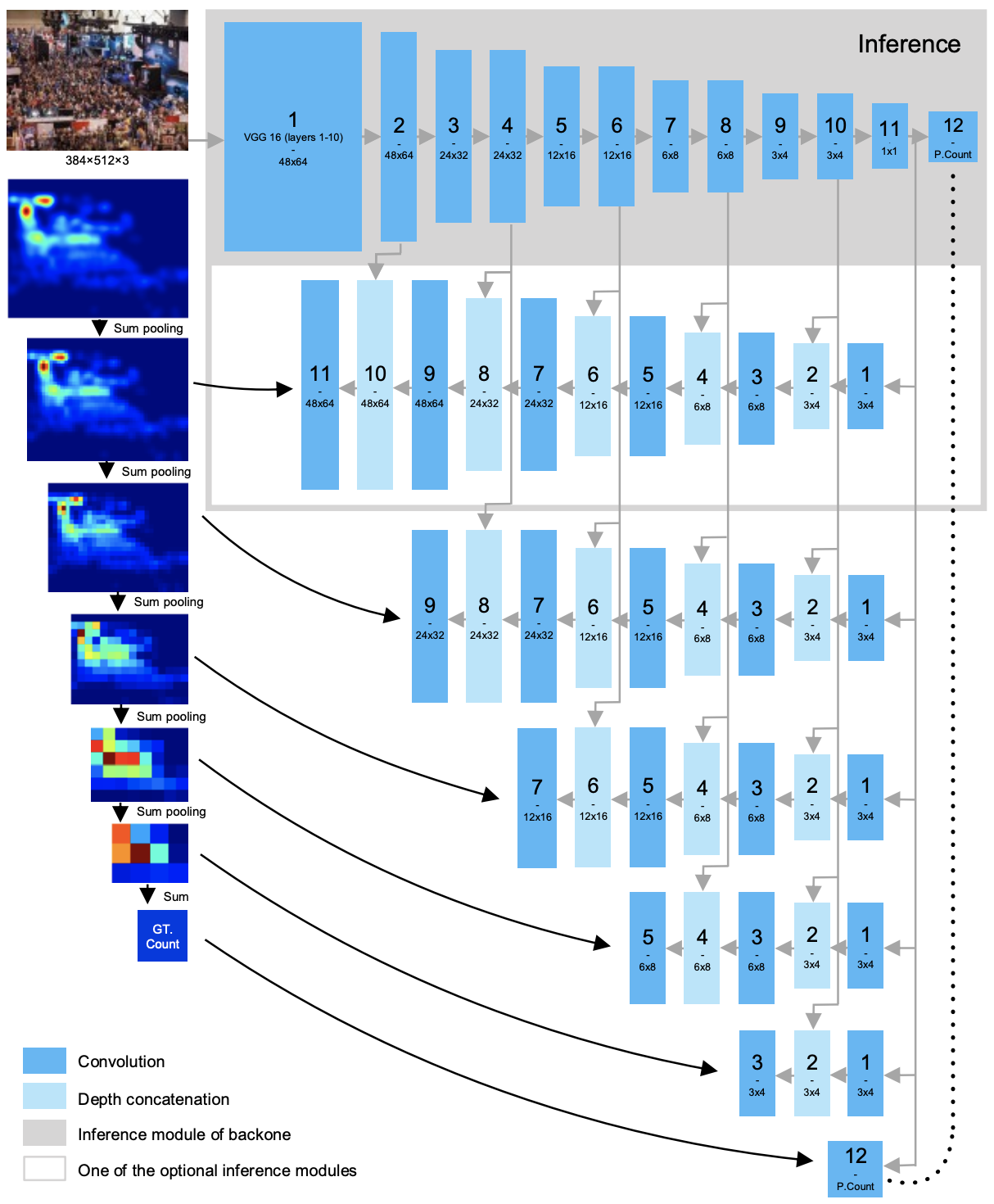}}
\caption{The architecture of the proposed DeepCount Model. Module in the grey box is the backbone regressor, below which are 5 branches predicting density maps. Large-size numbers on the blocks are 
referred to in detailed configuration in Table 1, while numbers in smaller font indicate the feature map dimensions.} \label{figure2}
\end{figure}

\newcommand{\tabincell}[2]{\begin{tabular}{@{}#1@{}}#2\end{tabular}}
\begin{table*}[t]
\begin{center}
{\caption{Configuration of DeepCount network. In the table, Conv and Conv-tr mean convolution and transposed convolution respectively. The pattern ${H\times W\times C\times C’}$ represents the dimension of convolution kernel. S denotes strides.}\label{table1}}
\resizebox{\textwidth}{!}{
\begin{tabular}{lcccccc}
\hline
\rule{0pt}{12pt}
Module&Backbone&Branch 1&Branch 2&Branch 3&Branch 4&Branch 5\\
\hline
\quad Input&\tabincell{c}{Image\\$384\times512\times3$}&$1\times1024$&$1\times1024$&$1\times1024$&$1\times1024$&$1\times1024$\\
\hline
\quad 1&\tabincell{c}{VGG 16\\Layers 1-10}&\tabincell{c}{Conv-tr-S1\\$3\times4\times1024\times256$}&\tabincell{c}{Conv-tr-S1\\$3\times4\times1024\times256$}&
	\tabincell{c}{Conv-tr-S1\\$3\times4\times1024\times256$}&\tabincell{c}{Conv-tr-S1\\$3\times4\times1024\times256$}&\tabincell{c}{Conv-tr-S1\\$3\times4\times1024\times256$}\\
\hline
\quad 2&\tabincell{c}{Conv-S1\\$3\times3\times512\times256$}&\tabincell{c}{Depth\\Concatenation}&\tabincell{c}{Depth\\Concatenation}&
	\tabincell{c}{Depth\\Concatenation}&\tabincell{c}{Depth\\Concatenation}&\tabincell{c}{Depth\\Concatenation}\\
\hline
\quad 3&\tabincell{c}{Conv-S2\\$3\times3\times256\times512$}&\tabincell{c}{Conv-tr-S2\\$4\times4\times512\times256$}&\tabincell{c}{Conv-tr-S2\\$4\times4\times512\times256$}&
	\tabincell{c}{Conv-tr-S2\\$4\times4\times512\times256$}&\tabincell{c}{Conv-tr-S2\\$4\times4\times512\times256$}&\tabincell{c}{Conv-S1\\$1\times1\times512\times1$}\\
\hline
\quad 4&\tabincell{c}{Conv-S1\\$3\times3\times512\times256$}&\tabincell{c}{Depth\\Concatenation}&\tabincell{c}{Depth\\Concatenation}&
	\tabincell{c}{Depth\\Concatenation}&\tabincell{c}{Depth\\Concatenation}\\
\hline
\quad 5&\tabincell{c}{Conv-S2\\$3\times3\times256\times512$}&\tabincell{c}{Conv-tr-S2\\$4\times4\times512\times256$}&\tabincell{c}{Conv-tr-S2\\$4\times4\times512\times256$}&
	\tabincell{c}{Conv-tr-S2\\$4\times4\times512\times256$}&\tabincell{c}{Conv-S1\\$1\times1\times512\times1$}\\
\hline
\quad 6&\tabincell{c}{Conv-S1\\$3\times3\times512\times256$}&\tabincell{c}{Depth\\Concatenation}&\tabincell{c}{Depth\\Concatenation}&
	\tabincell{c}{Depth\\Concatenation}\\
\hline
\quad 7&\tabincell{c}{Conv-S2\\$3\times3\times256\times512$}&\tabincell{c}{Conv-tr-S2\\$4\times4\times512\times256$}&\tabincell{c}{Conv-tr-S2\\$4\times4\times512\times256$}&
	\tabincell{c}{Conv-S1\\$1\times1\times512\times1$}\\
\hline
\quad 8&\tabincell{c}{Conv-S1\\$3\times3\times512\times256$}&\tabincell{c}{Depth\\Concatenation}&\tabincell{c}{Depth\\Concatenation}\\
\hline
\quad 9&\tabincell{c}{Conv-S2\\$3\times3\times256\times512$}&\tabincell{c}{Conv-tr-S2\\$4\times4\times512\times256$}&\tabincell{c}{Conv-S1\\$1\times1\times512\times1$}\\
\hline
\quad 10&\tabincell{c}{Conv-S1\\$3\times3\times512\times256$}&\tabincell{c}{Depth\\Concatenation}\\
\hline
\quad 11&\tabincell{c}{Conv-S1\\$3\times4\times256\times1024$}&\tabincell{c}{Conv-S1\\$1\times1\times512\times1$}\\
\hline
\quad 12&\tabincell{c}{Fc\\$1024\times1$}\\
\hline
\quad Output&P. Count&\tabincell{c}{P. Density Map\\$48\times64$}&\tabincell{c}{P. Density Map\\$24\times32$}&\tabincell{c}{P. Density Map\\$12\times16$}
	&\tabincell{c}{P. Density Map\\$6\times8$}&\tabincell{c}{P. Density Map\\$3\times4$}\\
\hline
\end{tabular}}
\end{center}
\end{table*}

\section{DeepCount}
\subsection{Gradient Fusion}
We regard our methodology of designing the network as Gradient Fusion. Multi-column methods such as MCNN\cite{zhang2016single} and CP-CNN\cite{sindagi2017generating} are feature fusion methods assembling different columns features from which are fused and gradients to which are separated. Fusing feature maps of multiple columns entails lots of computation overhead since each column cannot be without in order to make prediction. In contrast, the method of gradient fusion fuses only gradient matrices 
in backpropagation during training. We acknowledge that similar mechanisms are implemented in many works where they may be called otherwise. We use the name Gradient Fusion here to serve descriptive purpose to distinguish our methodology 
from other multi-column methods on crowd counting.

In our network configuration (see section 3.3), the backbone module is mostly shared with branches which produce different density maps of different scales. Multi-source gradients are fused together to train this critical backbone. Complexities of solving 
location estimation still exists for optimization of the parameters of the backbone. This restricts the backbone to learn location information to help branches predict density maps. This is a process which instils density-awareness to the backbone.

\subsection{Density map pixelation and improved SNR}
It takes great effort to label ground truth coordinates for heads in crowded images. Even with extra focus, getting wrong is inevitable. This brings error to the ground truth density maps. The correct density value of any data point in an 
original M$\times$N density map (corresponding to an M$\times$N image) is the signal of the data. We assume signal ($S$) is normally distributed and has non-zero mean $\mu_0$ ($S\sim\mathcal{N}(\mu_0,\,\phi)$).
We also assume the noise ($\epsilon$) a normally distributed random variable. It has zero mean, and standard deviation $\sigma_0$ ($\epsilon\sim\mathcal{N}(0,\,\sigma_0^2)$).
Density maps of different scales are produced by pixelating the original M$\times$N density map. Each pixelation operation is a sum pooling layer with both stride and window size equal 2. After $n$ levels of pixelation, a data point in the density map has signal mean
\begin{equation}
\mu_n=4^n\mu_0
\end{equation}
, and noise variance
\begin{equation}
\sigma_n^2=4^n\sigma_0^2.
\end{equation}
Therefore, signal-to-noise ratio (SNR) of the density map at the $n_{th}$ pixelation level is computed as
\begin{equation}
SNR_n=\frac{\mu_n^2}{\sigma_n^2}=C\cdot4^n,
\end{equation}
where $C$ is some constant. Thus, SNR of the training data in terms of every single data point grows exponentially for every pixelation operation, which means smaller density maps give more accurate information about the counting. But, on the other hand, some of the location information is lost after pixelation. Our network design is to maximum advantage of pixelated density map, and at the same time avoid any location information loss.

\subsection{Network configuration}
In order to signify our central point by comparing our model to others, we demonstrate a relatively simple network design in this paper. As shown in Figure 2, our proposed model consists of a straightforward down-sampling backbone and five branches interconnected to it. The backbone by itself has relatively low complexity. It functions as a deep CNN regressor which takes the crowd image as input and predicts the global count by regression. We design the network to have input size of $384\times512$ to cater most aspect ratios in practical uses, whereas arbitrary larger input image sizes are tackled by division and combination. Correspondingly, there are density maps of sizes \{${48\times64,24\times32,12\times16,6\times8,3\times4}$\} produced.

Specifically, the backbone has a frontend which extracts features from the input image. We transplant the first ten convolution layers from pretrained VGG-16 as our frontend model similar to CSRNet\cite{li2018csrnet}. The frontend produces feature maps of 8 times smaller spatial width and height relative to the input. Following are some $3\times3$ convolution layers to further dwindle the size of feature maps until when its spatial dimension matches the input dimension of a $3\times4$ convolution layer entering to produce a $1\times1024$ vector. We use $3\times3$ convolution with strides of 2 to halve the spatial dimension of the feature maps in the backend. In addition, a standard $3\times3$ convolution layer is put between two down-sampling layers to further deepen the network and to smooth the reduction of features. 

As for branches, they work in an up-sampling manner. Branches stemming from the last feature layer ($1\times1024$) of the backbone use transposed convolutions to up-sample the feature maps. To the output of each transposed convolution layer the deeper feature maps from backbone with the same dimension are channel-wise appended. Together, they form the input to the next transposed convolution layer. $1\times1$ convolution is used to reconstruct channels to produce density map prediction. At the end of the backbone, a scalar value is produced as the global count prediction. We call our network DeepCount in short for deep CNN count regressor. Table 1 details the network configuration.

\subsection{Generating ground truth density maps}
To produce ground truth density maps for training, we first apply convolution by fixed Gaussian kernel with standard deviation $\sigma$ = 5 (on the contrary of geometry-adaptive kernel adopted in most works) to generate density map of the same resolution as the original image, before sum pooling is employed to produce different levels of pixelated density maps. Five density maps are produced. Element sum of the density map is the ground truth count of the image.

\subsection{Objective function}
Labelling congested crowd data is indeed a painstaking task for human annotators. In some highly congested cases where the factual number of people is inevitably untraceable. This results in many annotations in congestions themselves being estimations and the ground truths becomes noisy. Also, for large scale heads, the annotation position is difficult to be exact. Hence, L1-norm loss is adopted to enhance robustness against noise as well as to convey steady updates to the network. We first define our objective function as:
\begin{equation}
L(\Theta) = \frac{1}{2N}\sum_{n=1}^N\sum_{k=1}^K\sum_{i,j}|y’_{nk} - f(X_{n},\Theta_{k})|_{ij}
\end{equation}
where $N$ is the size of the training batch, $K(K=6\ \&\ k\in\{1,2,...,6\})$ enumerates outputs of all branches and the global count regressor, $y’_{nk}$ is the ground truth density map (or the global count when $k=6$), $X_{n}$ is the input image and $\Theta_{k}$ denotes all parameters in model $f$ that contribute to making the corresponding $k_{th}$ prediction.

Given this objective function as basis, we add a multiplier $\beta$ to accentuate the importance of the global count prediction on backbone (where $k=6$). We notate it as a function of $k$:
\begin{equation}
B(k)=
\begin{cases}
\beta,& k=6\\
1,& k\neq6
\end{cases}
\end{equation}

Moreover, we add another hyperparameter $\omega$ to approximately adjust the loss to a reasonably small value ($<$10). An L2 regularisation term is also added to the function in an attempt to reduce overfitting. Hence, the objective function finally becomes:
\begin{equation}
L(\Theta) = [\frac{1}{2N}\sum_{n=1}^N\sum_{k=1}^K\sum_{i,j}|y’_{nk} - f(X_{n},\Theta_{k})|_{ij} \cdot B(k)] \cdot \omega + \frac{\lambda}{2}\|\Theta\|^2_{2}
\end{equation}

\subsection{Implementation}
VGG-16 model pretrained on ImageNet is used to initialise the frontend of the backbone. Therefore, input images are normalised in the same manner as how the VGG-16 model is trained. As for initialising the remaining part of the model, we use Xavier \cite{glorot2010understanding} initialisation for weights and a constant value 0 for biases. With the exception of the VGG-16 frontend where ReLU is the activation function, we set parametric ReLUs (leaky ReLU):
\begin{equation}
a(x)=
\begin{cases}
x,& x>0\\
\alpha x,& x\leq0
\end{cases}
\end{equation}
following every convolution layer. We sweep hyperparameters and choose ${\lambda=1\times10^{-5}, \omega=1\times10^{-2}}$ and $\beta=16$
for the objective function in equation (6), and $\alpha=0.2$ for the activation parameter \cite{xu2015empirical} in equation (7). We use Gradient Descent optimisation with momentum 0.9 and initial learning rate $1\times10^{-4}$ to train our model, except for pretrained parameters in frontend where learning rate is divided by a factor of 2 to initially $5\times10^{-5}$, Batch size $N$ is set to 32. On benchmark datasets, we train the network for around a hundred epochs. 

As alluded to above, to cope with images of varied sizes, we divide the original image to $384\times512$ crops to feed into our network. In inference, results from cropped images are to be merged to assemble the original ones again.

\section{Evaluation}
In this section, we report evaluation results yielded by our method introduced above. We evaluate our DeepCount network on four different public datasets: Shanghai Tech Part A and Part B \cite{zhang2016single}, UCF-QNRF \cite{idrees2018composition} and Mall \cite{chen2012feature}. Training details for all datasets are the same as mentioned in implementation section (section 3.6). In order to make fair comparison with benchmark results, we do no more data augmentation than random cropping and mirroring during training.
\subsection{Evaluation metrics}
For evaluation, we compute mean-absolute error (MAE) and root-mean-squared error (RMSE):
\begin{equation}
MAE = \frac{1}{N}\sum_{i=1}^N|C_{i} - C^{GT}_{i}|
\end{equation}
\begin{equation}
RMSE = \sqrt{\frac{1}{N}\sum_{i=1}^N||C_{i} - C^{GT}_{i}||^2}
\end{equation}
where $N$ is the number of testing images, $C_i$ and $C_i^{GT}$meaning predicted count and ground truth count respectively.

\subsection{Shanghai Tech}
Shanghai Tech\cite{zhang2016single} dataset includes Part A and Part B. Part A is the dataset for congested crowd counting. It has 241,677 annotations in 300 training images and 182 testing images with an average number 501. On the other hand, images in Part B are relatively sparse and all taken from streets in Shanghai. Our DeepCount model achieves state-of-the-art performance on both datasets. Test results are shown in Table 2.
\begin{table}
\setlength{\tabcolsep}{9pt}
\begin{center}
{\caption{Test results on Shanghai Tech Part A and Part B.}\label{table2}}
\begin{tabular}{lcccc}
\hline
\rule{0pt}{12pt}
&\multicolumn{2}{c}{Part A}&\multicolumn{2}{c}{Part B}\\
\hline
\\[-6pt]
\quad Method&MAE&RMSE&MAE&RMSE\\
\quad MCNN \cite{zhang2016single}&110.2&173.2&26.4&41.3\\
\quad Switching CNN \cite{babu2017switching}&90.4&135.0&21.6&33.4\\
\quad DecideNet \cite{liu2018decidenet}&-&-&20.75&29.42\\
\quad CP-CNN  \cite{sindagi2017generating}&73.6&106.4&20.1&30.1\\
\quad ic-CNN \cite{ranjan2018iterative}&68.5&116.2&10.7&16.0\\
\quad CSRNet \cite{li2018csrnet}&68.2&115.0&10.6&16.0\\
\quad PSDDN+ \cite{liu2019point}&65.9&112.3&9.1&14.2\\
\quad SANet \cite{cao2018scale}&67.0&\textbf{104.5}&8.4&13.6\\
\quad DeepCount (ours)&\textbf{65.2}&112.5&\textbf{7.2}&\textbf{11.3}
\\
\hline
\end{tabular}
\end{center}
\end{table}

\subsection{UCF-QNRF}
The UCF-QNRF \cite{idrees2018composition} dataset has a greater number of annotations (1,251,642) in higher quality images of a wider variety of scenes, including sparse and dense ones. There are extremely dense scenes in this dataset, so much so that a single image may have maximumly 12,865 of annotations in UCF-QNRF. It is considered a harder one. Our method outperforms current methods (see Table 4).

\begin{table*}[t]
	\begin{center}
		{\caption{Comparing between configurations and FLOPs of CSRNet and our DeepCount. Branch one (middle column) predicts the same- size density maps as does CSRNet, while backbone predicts a global count without producing any density maps.}\label{table3}}
		\resizebox{\textwidth}{!}{
			\begin{tabular}{ccc|ccc|ccc}
				\hline
				\multicolumn{3}{c}{CSRNet (backend)}&\multicolumn{3}{c}{DeepCount Branch 1}&\multicolumn{3}{c}{DeepCount (backend))}\\
				\hline
				\quad Layer&Output&\tabincell{c}{Million\\FLOPs}&Layer&Output&\tabincell{c}{Million\\FLOPs}&Layer&Output&\tabincell{c}{Million\\FLOPs}\\
				\hline
				\quad \tabincell{c}{Conv-s1\\$3\times3\times512\times512$}&$48\times64\times512$&7248&\tabincell{c}{Conv-tr-s1\\$3\times4\times1024\times256$}&$3\times4\times256$&37
				&\tabincell{c}{Conv-s1\\$3\times3\times512\times256$}&$48\times64\times256$&3624\\
				\hline
				\quad \tabincell{c}{Conv-s1\\$3\times3\times512\times512$}&$48\times64\times512$&7248&\tabincell{c}{Conv-tr-s2\\$4\times4\times512\times256$}&$6\times8\times256$&101
				&\tabincell{c}{Conv-s2\\$3\times3\times256\times512$}&$24\times32\times512$&906\\
				\hline
				\quad \tabincell{c}{Conv-s1\\$3\times3\times512\times512$}&$48\times64\times512$&7248&\tabincell{c}{Conv-tr-s2\\$4\times4\times512\times256$}&$12\times16\times256$&403
				&\tabincell{c}{Conv-s1\\$3\times3\times512\times256$}&$24\times32\times256$&906\\
				\hline
				\quad \tabincell{c}{Conv-s1\\$3\times3\times512\times256$}&$48\times64\times256$&3624&\tabincell{c}{Conv-tr-s2\\$4\times4\times512\times256$}&$24\times32\times256$&1611
				&\tabincell{c}{Conv-s2\\$3\times3\times256\times512$}&$12\times16\times512$&226\\
				\hline
				\quad \tabincell{c}{Conv-s1\\$3\times3\times256\times128$}&$48\times64\times128$&906&\tabincell{c}{Conv-tr-s2\\$4\times4\times512\times256$}&$48\times64\times256$&6442
				&\tabincell{c}{Conv-s1\\$3\times3\times512\times256$}&$12\times16\times256$&226\\
				\hline
				\quad \tabincell{c}{Conv-s1\\$3\times3\times128\times64$}&$48\times64\times64$&226&\tabincell{c}{Conv-s1-p0\\$1\times1\times512\times1$}&$48\times64\times1$&2
				&\tabincell{c}{Conv-s2\\$3\times3\times256\times512$}&$6\times8\times512$&57\\
				\hline
				\quad \tabincell{c}{Conv-s1\\$1\times1\times64\times1$}&$48\times64\times1$&0.2&&&&\tabincell{c}{Conv-s1\\$3\times3\times512\times256$}&$6\times8\times256$&57\\
				\hline
				\quad &&&&&&\tabincell{c}{Conv-s2\\$3\times3\times256\times512$}&$3\times4\times512$&14\\
				\hline
				\quad &&&&&&\tabincell{c}{Conv-s1\\$3\times3\times512\times256$}&$3\times4\times256$&14\\
				\hline
				\quad &&&&&&\tabincell{c}{Conv-s1-p0\\$3\times4\times256\times1024$}&$1\times1\times1024$&3\\
				\hline
				\quad &&&&&&\tabincell{c}{Conv-s1-p0(Fc)\\$1\times1\times1024\times1$}&1&0.001\\
				\hline
				\quad Total&&26500&&&8596&&&6034
				\\
				\hline
		\end{tabular}}
	\end{center}
\end{table*}

\begin{table}[htb]
	\setlength{\tabcolsep}{20pt}
	\begin{center}
		{\caption{Test results on UCF-QNRF.}\label{table4}}
		\begin{tabular}{lcc}
			\hline
			\rule{0pt}{12pt}
			Method&\multicolumn{1}{c}{MAE}&\multicolumn{1}{c}{RMSE}\\
			\hline
			\\[-6pt]
			\quad Idrees et al.(2013) \cite{idrees2013multi}&315&508\\
			\quad MCNN \cite{zhang2016single}&277&426\\
			\quad CMTL \cite{sindagi2017cnn}&252&514\\
			\quad Switching CNN \cite{babu2017switching}&228&445\\
			\quad Resnet101 \cite{he2016deep}&190&277\\
			\quad Densenet201 \cite{huang2017densely}&163&226\\
			\quad Idrees et al.(2018) \cite{idrees2018composition}&132&191\\
			\quad TEDnet \cite{jiang2019crowd}&113&188\\
			\quad DeepCount (ours)&\textbf{95.7}&\textbf{167.1}
			\\
			\hline
		\end{tabular}
	\end{center}
\end{table}

\begin{table}[htb]
\setlength{\tabcolsep}{20pt}
\begin{center}
{\caption{Test results on Mall.}\label{table5}}
\begin{tabular}{lcc}
\hline
\rule{0pt}{12pt}
Method&\multicolumn{1}{c}{MAE}&\multicolumn{1}{c}{RMSE}\\
\hline
\\[-6pt]
\quad R-FCN \cite{dai2016r}&6.02&5.46\\
\quad Faster R-CNN \cite{ren2015faster}&5.91&6.60\\
\quad COUNT Forest \cite{pham2015count}&4.40&2.40\\
\quad Weighted VLAD \cite{sheng2016crowd}&2.41&9.12\\
\quad DecideNet \cite{liu2018decidenet}&\textbf{1.52}&\textbf{1.90}\\
\quad DeepCount (ours)&1.55&2.00
\\
\hline
\end{tabular}
\end{center}
\end{table}

\subsection{Mall}
Unlike the above datasets, images from Mall dataset\cite{chen2012feature} are surveillance video frames from a static viewpoint at a same venue. There are 800 frames for training and the other 1200 for testing. Since crowds in the dataset are sparse, Mall is not as challenging as others. Although previous methods have shown very promising results on this dataset, we still evaluate our model on it to demonstrate its excellent performance on invariant scene and as well to make comparison with some detection-based methods. (see Table 5). 

\begin{figure*}[t]
	\centerline{\includegraphics[scale=0.5]{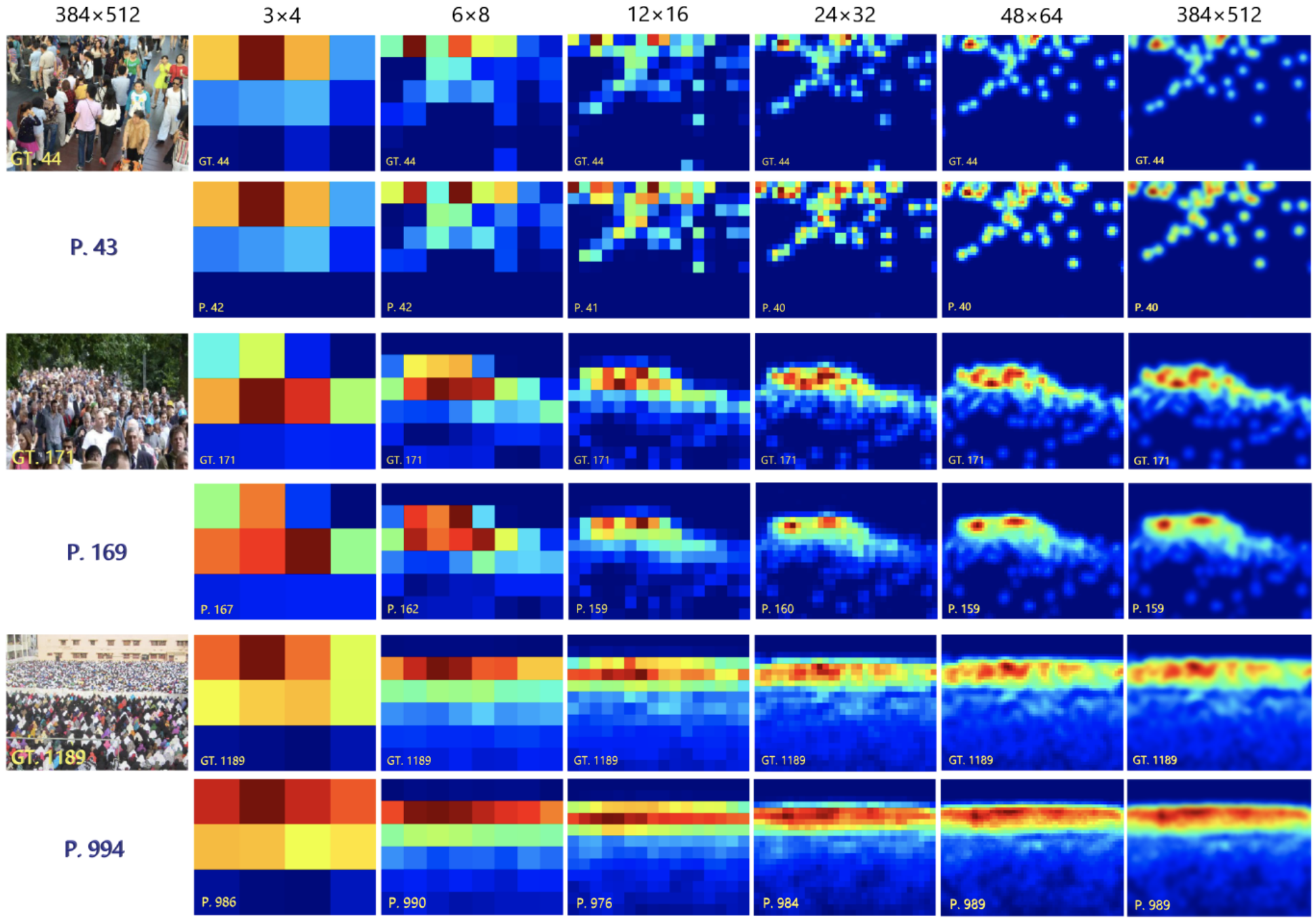}}
	\caption{Examples of ground truths and output density maps of test data from Shanghai Tech Part B (top), Part A (middle), UCF-QNRF (bottom). Density maps in the row starting with an image are the ground truths. The ones below are the predictions. The number in blue below the image is the predicted global count by backbone alone. The last column shows original size ground truth density maps and prediction density maps up-sampled from outputs of branch one, up-sampling being done by bilinear interpolation and separable Gaussian filter.} \label{procstructfig}
\end{figure*}

\section{Discussion}
\subsection{Capacity and velocity}
Arguably, the more parameters a neural network has, the greater its potential is to have high capacity to model the underlying relationship of the random variables. Although many cases suggest otherwise, we do often see positive correlation between extra parameters and increments of performance\cite{li2018csrnet,simonyan2014very,he2016deep,howard2017mobilenets}. Be that as it may, we still tend to avoid expensive computation a larger network would bear in practice. Trading off between capacity and velocity has been a dilemma for long. Hereby, we explicate how the idea of our DeepCount network is able to pursue both capacity and velocity at the same time by comparing it with CSRNet\cite{li2018csrnet} in whose paper Li et al. argue cogently about the effect of number of parameters and design efficiency. 

CSRNet and our backbone network use the same VGG-16 frontend. In the backend, CSRNet predicts the density map and our model predicts a global count. Assuming they receive the same $384\times512\times3$ input (backends thus receive $48\times64\times512$ input), we detail their layer configuration with corresponding output and computation cost of each layer in Table 3. In addition, we add branch one which predicts the same density map as CSRNet to the table. Computation cost is calculated in terms of number of floating-point operations (FLOPs) that happens throughout a forward pass in the backend. Number of FLOPs of one convolution layer is computed as:
\begin{equation}
FLOPs=H\cdot W\cdot C\cdot K_1\cdot K_2\cdot C’
\end{equation}
where it depends multiplicatively upon output feature map size $H\times W$, convolution kernel size $K_1\times K_2$, number of output channels $C$, and number of input channels $C’$.

\begin{table}[t]
	\begin{center}
		{\caption{Comparing number of parameters and inference speed between CSRNet and the backbone of our DeepCount model.}\label{table6}}
		\begin{tabular}{lcccc}
			\hline
			\rule{0pt}{12pt}
			Method&\tabincell{c}{SHT Part B\\MAE}&\tabincell{c}{Million\\Parameters}&FPS&Speedup\\
			\hline
			\\[-6pt]
			\quad CSRNet&10.6&16.3&33&$1\times$ \\
			\quad DeepCount(Ours)&7.2&\tabincell{c}{21.4\\(58.1 in total)}&45&$1.4\times$ 
			\\
			\hline
		\end{tabular}
	\end{center}
\end{table}
We also measure number of parameters as well as frame-per-second (FPS) for both networks (see Table 6). Run time evaluation is performed on one NVIDIA Tesla P40 GPU. As mentioned above, the backbone of our DeepCount model can be a standalone network detached from the rest in inference, and thereby becomes a count regressor without computing the computationally expensive density maps, and noticeably with better performance compared to other approaches. 

As shown in Table 3 and Table 6, having a deeper architecture and greater preponderance of parameters (58.1 million for training and 21.4 million for inference) though, our DeepCount backbone does count inferences with much less FLOPs and therefore in higher velocity, and perhaps more importantly, with higher accuracy. These quantitative results suggest our proposed DeepCount model has the ability of accommodating more variations while making faster and better prediction. This implies its nature of outstanding capacity and efficiency. 

\subsection{Comparison on branches}
\begin{table*}[t]
\setlength{\tabcolsep}{16pt}
\begin{center}
{\caption{Comparing between outputs on different branches.}\label{table7}}
\begin{tabular}{lcccccc}
\hline
\rule{0pt}{12pt}
&\multicolumn{1}{c}{Branch 1}&\multicolumn{1}{c}{Branch 2}&\multicolumn{1}{c}{Branch 3}&\multicolumn{1}{c}{Branch 4}&\multicolumn{1}{c}{Branch 5}&\multicolumn{1}{c}{Backbone}\\
\hline
\\[-6pt]
\quad Output Size&$48\times64$&$24\times32$&$12\times16$&$6\times8$&$3\times4$&$1\times1$ \\
\quad Million FLOPs&6034&2152&541&138&38&-\\
\quad Shanghai Tech Part A&79.2&73.4&69.7&66.7&65.8&65.2\\
\quad Shanghai Tech Part B&9.7&8.9&8.0&7.4&7.2&7.2\\
\quad UCF-QNRF&193.3&185.0&155.3&112.3&96.9&95.7\\
\quad Mall&4.74&2.89&2.46&1.77&1.56&1.55
\\
\hline
\end{tabular}
\end{center}
\end{table*}

As we have obtained the global count regressed by backbone, we can as well integrate the output density map to make count prediction like common density map-based methods. In the following, we compare predictions made between branches on MAE. Besides, we compute FLOPs for each branch to analyse their computational costs. Results are shown in Table 7.

As shown, the larger the density map, the harder it is to be precise. Immediate reasons for this may be that larger density maps are sparser and usually awash with noise caused mostly by annotations of large-scale heads. Our method avoids predicting the count relying merely on density maps but exploits useful information from them to rather train the global count regressor. This allows more accurate predictions to be achieved. 

Despite extra computation, there are situations in which density map, which gives extra information about the distribution, becomes a requirement. Our model can make directly count inference with backbone at its full speed while optionally producing density maps of multiple resolutions. User can choose smaller density maps to reduce computational expensiveness or larger ones to get more illuminating impression about the crowd distribution. Using bilinear interpolation and separable Gaussian filter, the largest density map produced can be efficiently up-sampled to original resolution for high-definition display, thus we argue it is unworthy to train a network to produce a high-resolution density map. Figure 3 shows our predicted density maps compared to their ground truths.

\subsection{Significance of gradients}
Gradients are considered crucial to the achievement of our model. Hence, we detail more experiments to further cast light on the importance of them. 

Since derivatives of ReLU are a staircase function suppressing the negative direction, gradients in half of its activation space are set to naught. The back-propagation gradient matrices are sparse and may hinder propagation of gradient flow and counterproductively cause a large part of the network underused. Instead, Parametric ReLU has non-zero gradients in all quadrants allowing the network to fully learn. Table 8 shows results of training a network with all ReLU activations in comparison with our baseline network. As shown in Table 8, when gradients are sparse, the capability of the network drops.
\begin{table}
\setlength{\tabcolsep}{37pt}
\begin{center}
{\caption{Comparing results between using ReLU(sparse gradients) and PReLU(full gradients).}\label{table8}}
\begin{tabular}{lc}
\hline
\rule{0pt}{12pt}
Activation&SHT Part B MAE\\
\hline
\\[-6pt]
\quad ReLU&8.3\\
\quad PReLU&7.2
\\
\hline
\end{tabular}
\end{center}
\end{table}

Also, the idea of being density-aware by gradient fusion is to leverage gradients sourced from supervision of multi-scale density maps. In ablation experiments (see Table 9), as we detach branches one by one from the largest to the smallest in training, the backbone receives less gradients in each case, and then the trend of performance degradation becomes more and more apparent. We also find that the model results satisfactorily when even trained with only global count on a dense dataset, while on a difficult one (sparse, non-uniform) it does not converge always, but we see it resolve if pretrained on dense dataset.
\begin{table}
\setlength{\tabcolsep}{30pt}
\begin{center}
{\caption{Ablation on branches.}\label{table9}}
\begin{tabular}{lc}
\hline
\rule{0pt}{12pt}
Ablation&SHT Part B MAE\\
\hline
\\[-6pt]
\quad No ablation&7.2\\
\quad Branch 1 ablated&7.4\\
\quad Branch 1-2 ablated&7.7\\
\quad Branch 1-3 ablated&8.2\\
\quad Branch 1-4 ablated&8.3\\
\quad Branch 1-5 ablated&9.1
\\
\hline
\end{tabular}
\end{center}
\end{table}

By means of this, we are now safer to conclude that the abundance of gradients has advantageous influence on our network and parameters in branches are indeed instrumental in the training of backbone. Giant as it may be, the network of branches is not a concern in an inference deployment. Unless training efficiency is also in a serious consideration, having a rationally greater number of parameters in this auxiliary module should be deemed innocuous as long as performance does not remain stagnant.

\section{Conclusion}
In this paper, we discuss advantages and limitations of current crowd counting methods, in light of which we propose a novel DeepCount network to be both fast and precise on count prediction and flexible on density map generation. State-of-the-art performance on public datasets evidences the effectiveness of our method. Our code is publicly available at:

https://github.com/GeorgeChenZJ/deepcount

\ack We would like to thank the anonymous reviewers for their thoughtful
reading and comments.

\bibliography{ecai}

\end{document}